\pdfoutput=1
\documentclass{article}

     \usepackage[preprint,nonatbib]{neurips_2018}

\usepackage[utf8]{inputenc} % allow utf-8 input
\usepackage[T1]{fontenc}    % use 8-bit T1 fonts
\usepackage{caption}
\usepackage{hyperref}       % hyperlinks
\usepackage{url}            % simple URL typesetting
\usepackage{booktabs}       % professional-quality tables
\usepackage{amsfonts}       % blackboard math symbols
\usepackage{nicefrac}       % compact symbols for 1/2, etc.
\usepackage{microtype}      % microtypography
\usepackage{graphicx}
\usepackage{subcaption}
\usepackage{mwe}
\usepackage{verbatim} 
\usepackage{amsmath}

\title{Don't ignore Dropout in Fully Convolutional Networks}

\author{%
  Thomas Spilsbury \\
  Aalto University \\
  Espoo, Finland \\
  \texttt{thomas.spilsbury@aalto.fi} \\
  \And
  Paavo Camps \\
  Aalto Univeristy \\
  Espoo, Finland \\
  \texttt{paavo.camps@aalto.fi} \\
}

\begin{document}

\maketitle

\begin{abstract}
   Data for image segmentation models can be costly to obtain due to the precision required by human annotators. We run a series of experiments showing the effect of different kinds of Dropout training on the DeepLabv3+ image segmentation model when trained using a small dataset. We find that when appropriate forms of Dropout are applied in the right place in the model architecture that non-insignificant improvement in Mean Intersection over Union (mIoU) score can be observed. In our best case, we find that applying Dropout scheduling in conjunction with \texttt{SpatialDropout} improves baseline mIoU from 0.49 to 0.59. This result shows that even where a model architecture makes extensive use of Batch Normalization, Dropout can still be an effective way of improving performance in low data situations.
\end{abstract}

\section{Introduction}

In recent years, the use of vanilla Dropout on the convolutional layers of Convolutional Neural Networks \cite{journals/corr/HintonVD15} has been called into question by researchers \cite{conf/nips/GhiasiLL18} \cite{conf/cvpr/TompsonGJLB15}. The main problem arises from the fact that activations along the convolutional layers are spatially correlated within each generated feature map, meaning that the use of random dropout across each pixel leads only to an effective learning rate reduction and does not have any regularizing effect \cite{conf/cvpr/TompsonGJLB15}. In addition, the introduction of Batch Normalization \cite{conf/icml/IoffeS15}, has also reduced the need for a solution like dropout, because \texttt{BatchNorm} has regularizing properties of its own. Indeed, there is also theoretical justification to show that the combination of \texttt{BatchNorm} and Dropout may have \emph{harmful} results due to the variance shift introduced at test time when Dropout is disabled \cite{journals/corr/abs-1801-05134}.

We show, through some simple experiments on the image semantic segmentation problem with limited training data, that carefully applied Dropout can still be an effective mechanism for reducing over-fitting and improving test-time accuracy, and therefore should be considered by practitioners when the amount of training data available for a problem is small.

The Semantic Segmentation problem is an open optimization problem in Computer Vision. Given some image $I$, the task is to find
some \emph{segmentation} $S$, such that each output pixel $S_{ij}$ corresponds to some known class at the correct position in the image. The goal is to maximize over some unseen test set, the \emph{Mean Intersection Over Union (mIoU)} between the true segmentation $S'$ and the output segmentation $S$.

$$
\texttt{mIoU} = \frac{S' \cap S}{S' \cup S} = \frac{1}{mn} \sum_{mn} \frac{S_{ij} = S_{ij}}{S_{ij} \ne S_{ij}} 
$$

The authors of the current state-of-the-art method\footnote{On PASCAL VOC2012, as at 12 May 2019 on the paperswithcode.com leaderboard} for semantic image segmentation (\texttt{DeepLabV3+}) make no mention of Dropout based regularization in their paper \cite{journals/corr/abs-1802-02611}, possibly due to the problems illustrated with Dropout above.

We reimplemented this image segmentation architecture and provide empirical results showing that application of different forms of Dropout (such as \texttt{SpatialDropout} \cite{conf/cvpr/TompsonGJLB15}, \texttt{DropBlock} \cite{conf/nips/GhiasiLL18} and \texttt{UOut} \cite{journals/corr/abs-1801-05134}) can provide an mIoU boost on validation data when the training data is limited. This is an important problem to focus on, since obtaining training data for semantic image segmentation is considerably more difficult than for other tasks such as object detection and image classification, as it requires precise labelled data from human annotators.

\section{Model Architecture}

A brief discussion of the design and justification for the \texttt{DeepLabV3+} architecture follows.

If we were to break down the image segmentation task into its components, we see that any proposed architecture would need to do the following:

\begin{itemize}
    \item Detect features in the image.
    \item Group detected features into segmented objects.
    \item Decode the grouping encoding into a pixelwise segmented Image.
\end{itemize}

Then, on a high level, the \texttt{DeepLabV3+} architecture achieves each of these objectives as follows:

\begin{itemize}
    \item \emph{Feature Detection}. This can be achieved using a sufficiently deep hierarchical feature detection architecture such as ResNet \cite{journals/corr/HeZRS15}, Xception \cite{Chol17Xception}. In our implementation, ResNet is used.
    \item \emph{Perceptual Grouping}. The grouping needs to be scale invariant, such that an architecture that learns weights required to make a perceptual grouping for an object at scale $S$ can make the same perceptual grouping at scale $S'$. This is achieved through a pyramid pooling architecture \cite{journals/corr/HeZR014}, where groups of features are highlighted at different scales in the image, then the result is linearly combined via $1 \times 1$ convolution.
    \item \emph{Decoding}. This can be achieved with achieved with a fully convolutional architecture that combines that low level features with the perceptual grouping features to produce a feature map for each class, with a given feature map having a stronger activation in $A^{(k)}_{ij}$ if it is more probable that class $k$ is present at co-ordinates $ij$ in the image.
\end{itemize}

In our experiments, we apply different Dropout based regularization at each stage to explore the overall effectiveness of the methods at each stage.

\section{Dropout Approaches}

\subsection{Vanilla Dropout}

Vanilla Dropout\cite{srivastava2014dropout}, sometimes referred to as \texttt{Dropout1D}, randomly sets each activation on the last dimension to zero, with probability $p$ at training time. The effect is that the following layers may not rely on the presence of a sparse number of activations on the layers below and instead must generalize by relying on all the features available to it. Srivastava \& Hinton show that such an approach is the equivalent of making inference using an ensemble of all possible subnetworks at that layer. At test time, all activations are scaled by ${1 - p}$ such that the mean value of each feature map is the same at both training and test time.

As illustrated in \cite{conf/cvpr/TompsonGJLB15} and \cite{conf/nips/GhiasiLL18}, such an approach lacks theoretical justification for feature maps produced by convolutional layers as the pixel-activations produced on each feature map in the final dimension are spatially correlated. This means that even if a single activation in a feature map is dropped, there is still a high probability that the feature will still be used by later layers by the corresponding pixel activations. The probability that that the entire feature will be dropped is actually $p_f = \prod^j p$, where $j$ is the number of activations making up the given feature. Where $j$ is large, then $p_f$ will be very low indeed. This drawback makes Vanilla Dropout ill-suited to convolutional networks.

\subsection{Channel Dropout}
In contrast, Channel Dropout (otherwise known as \texttt{SpatialDropout}) \cite{conf/cvpr/TompsonGJLB15} applies Dropout along the "channel" dimension of each activation tensor, setting entire feature maps to zero, with probability $p$.

Such an approach has a justification in the sense very large convolutional networks typically produce several hundred channels of feature maps at each layer, with each feature maps corresponding the presence of a given feature within an image in principle. Therefore, applying Dropout along the channels completely removes a given feature from the output within the entire Image, meaning that the upper layers must do higher level feature detection using a combination of all features available to them. One should observe, however, that due to stride-based downsampling applied by most architectures, network outputs are not equivariant to translation \cite{zhang2019shiftinvar}. It is therefore possible that a filter for a given channel may be detecting multiple features depending on their location in the image. Channel Dropout may remove features that are correlated with each other.

\subsection{DropBlock}
\texttt{DropBlock} introduced by \cite{conf/nips/GhiasiLL18} takes a similar approach, but instead applies dropout contiguous patches within an Image. The justification is similar to that in \cite{conf/cvpr/TompsonGJLB15} - pixelwise features within images are spatially correlated with each other, so more than a single pixel must be removed in order for Dropout to have any effect. In contrast to Channel Dropout, such an approach may not remove features that are correlated with each other, since only a part of each feature map is dropped. However, a drawback of DropBlock is that it partially reintroduces the problems encountered by Vanilla Dropout, in that the removed patch may only overlap with part of feature from an Image, thereby only resulting in an effective learning rate reduction, but no regularization. In addition, the gradient between the dropped area and the signal area of each channel will be very sharp, potentially introducing a substantial amount of noise.

\subsection{UOut}

Li, Xiang et al., showed that insertion of Dropout layers before Batch Normalization layers in a model
architecture may negatively affect the performance of the later Batch Normalization layers due to variance
shift that occurs when Dropout is not applied \cite{journals/corr/abs-1801-05134}. In effect, the Batch Normalization layers learn to counteract
covariate shift in the data that no longer exists when Dropout is turned off at test time. Specifically, the authors showed that if the inputs $x_i$ come from a distribution with $\mu$ mean and $v$ variance then the variance after applying dropout is given by $(\frac{1}{p}(\mu^2 + v) - \mu^2)(\sum w^2_i + \rho^{ax} \sum_i \sum_{j \ne i} w_i w_j)$, where $\rho^{ax}_{ij} = \frac{Cov(a_ix_i, a_jx_j)}{\sqrt{Var(a_ix_i)}\sqrt{Var(a_jx_j)}}$.
They then show that the variance at test time when Dropout is not applied is given by $v(\sum w^2_i + \rho^{x} \sum_{i} \sum_{j \ne i} w_j w_j)$. Then, the variance shift between train and test time is given by $\frac{v(\sum w^2_i + \rho^{x} \sum_{i} \sum_{j \ne i} w_j w_j}{(\frac{1}{p}(\mu^2 + v) - \mu^2)(\sum w^2_i + \rho^{ax} \sum_i \sum_{j \ne i} w_i w_j)}$. This simpifies to $\frac{v}{(\frac{1}{p}(\mu^2 + v) - \mu^2)}$, meaning that for instance, if the data was mean-centered and standard deviation scaled, and the dropout ratio was 0.1 the variance would be scaled by 0.9. The authors then propose \texttt{UOut}
to avoid this problem. In short, \texttt{UOut} adds uniform noise along one of the dimensions of the training data,
with distribution similar to that of the original dropout probability $x_i + x_ir_i, r_i \sim U[-\beta, \beta]$. In that case, the variance shift rate is given by $\frac{v}{E((x_i + x_ir_i)^2)}$. The authors show that comparably, where $\beta = 0.1$, then the variance shift will be $\frac{v}{v + 0.01}$, which for a normalized $v$ is much closer to 1 than 0.9. Therefore, variance shift with \texttt{UOut} will be much smaller than variance shift with any other dropout method.

In our implementation of \texttt{UOut}, we select one scalar value on the distribution $U[-p, p]$ for each channel, then
apply it to that channel.

\subsection{ScheduledDropPath}

Dropout only becomes necessary once the model is \emph{overfitting}, which typically
begins to happen later on in the training process. Applying Dropout early on in
the process may prevent the model from finding good local optima initially. To counteract this effect, the authors of \cite{conf/cvpr/ZophVSL18} suggest linearly increasing the dropout probability (for both \texttt{SpatialDropout},
\texttt{DropBlock} and \texttt{UOut} over a fixed ramp early in training until
a point where overfitting is likely to happen and Dropout should be most effective). This method is known as \texttt{ScheduledDropPath}. In this paper
separate experiments were performed using \texttt{ScheduledDropPath} with the
hyperparameter $n = 30$, such that each dropout method would be fully effective
at 30 epochs.

\section{Experimental Dataset}

We run our experiments on the standard benchmark for the image segmentation, the PASCAL Visual Object Classes Contest 2012 Semantic Segmentation dataset \cite{pascal-voc-2012}. The dataset contains 1464 images in the training set and 1449 images in the validation set.

Segmentation class labels are specially formatted \texttt{png} files where each pixel value
corresponds to a class label from $[0, 21)$. Since the difference in pixel values would normally
be imperceptible to a human viewer, these images are typically visualized by allocating a new
color to each class label, then visualizing the resulting Image.

\begin{table}[h]
\centering
\begin{minipage}{\textwidth}
\begin{tabular}{ll}
Training Transformations                                & Validation Transformations                                         \\
\hline
Cropping and Centering at 513 x 513 with random scaling \footnote{Between 0.5 and 2.0} & Cropping and Centering at 513 x 513 \\
Random Horizontal Flip                                  & image Normalization (with ImageNet weights)                        \\
Random Gaussian Blur (radius between 0.0 to 1.0)        &                                                                    \\
image Normalization (with ImageNet weights)             &                                    
\end{tabular}
\end{minipage}
\captionof{table}{Transformations}
\label{tab:transformations}
\end{table}

The fully convolutional architecture of the network means that in principle, an image of any
size can be provided as input to the network, however in practice the
"standard" data augmentation transformations given in Table \ref{tab:transformations} are used.

\section{Experimental Methods}

For the purposes of the experiments below, we keep 10\% of each class from the training set, then test using the entire validation set.
This ensures that we retain the same balance of classes within the training set, but that without regularization we would be likely to overfit the training data because there is so little of it.
For the purposes of this paper, this forced overfitting is motivated by the fact that with sufficient data, fitting deep models becomes a more
difficult problem than dealing with overfitting, but there may be scenarios where users of image segmentation architectures may need to
be data-efficient, as labelling data for the image segmentation task takes a large degree of human precision and effort.

\subsection{Training Process}

The training process for our proposed model follows the standard practice for how convolutional
neural networks are trained in general, however the following key differences should be observed:

\begin{itemize}
    \item \emph{Pre-training of the Feature Detection Layers}: We bootstrap the image segmentation process by
    using pre-trained weights on the feature detection network (obtained by pre-training on
    ImageNet). This is similar to the approach taken in the \texttt{DeepLabv3+} paper \cite{journals/corr/abs-1802-02611}. In general, pre-training is useful because it allows
    knowledge of low-level features from images in-general to be transferred to other tasks
    such as image segmentation without having to re-learn those low-level features from
    scratch \cite{journals/corr/HintonVD15}. We found that where pre-training was not used, we were
    never able to reach satisfactory results in the image segmentation tasks, with the mIoU metric
    only reaching at most 0.2 after training for 50 epochs, which is about one day's worth of training
    time.
    \item \emph{Discriminative Fine Tuning}: Since we already have knowledge of what low-level features
    of images in general are through the pre-training mechanism discussed above, we don't want to
    de-tune that knowledge during the training process for the segmentation layers. Therefore we apply
    a 10x lower learning rate to the feature detection network parameters as compared to the
    Spatial Pyramid Pooling and Decoder parameters which have not been trained. This was not discussed
    in the \texttt{DeepLabv3+} paper, though is present in the canonical PyTorch reimplementation\cite{jfzhang-repo}.
    Discriminative Fine Tuning was first discussed in the literature
    within the context of transfer learning in Natural Language Processing, but is applicable to
    transfer learning in any other context such as Computer Vision.
    \item \emph{Decreasing Polynomial Learning Rate Schedule}: This is a learning rate policy that sets
    the "base learning rate" (before differential learning rate multipliers are applied) to
    $(1 - \frac{I}{M})^{0.9}$ where $I$ is the current "iteration" given by
    $\text{epoch} \times \frac{\text{batch\_index}}{\text{num\_batches}}$ and $M$ is the maximum number of possible
    "iterations" $\text{epochs} \times \text{num\_batches}$. Empirically this has been shown to
    yield better results than fixed step-multiplier learning rates \cite{ChePap17}, though there does not
    appear to be any theoretical analysis in the literature as to why this is the case.
    \item \emph{Data Augmentation}: We apply data augmentation when loading training data as
    described in Table \ref{tab:transformations}, however as a special case the \emph{same} transformations
    need to be applied to \emph{both} the input image \emph{and} the segmentation label map in order to
    ensure that they remain aligned.
    \item \emph{Loss Function}: The loss function is a variant of standard Cross-Entropy loss between
    a sparse label encoding and a log-probability distribution given by the $\text{softmax}$ operator.
    To account for the fact that loss needs to be computed for each pixel in the segmentation map, the
    loss is computed by the taking vector along the \emph{channels} of the network output for each pixel
    and comparing it to the label for that pixel.
\end{itemize}

As for the other details of training, we use Stochastic Gradient Descent as the optimizer with
momentum of 0.9 and a weight decay of $5 \times 10^{-4}$ and a base learning rate of $7 \times 10^{-3}$
(such that the effective learning rate for the Decoder and Spatial Pyramid Pooling Parameters is
$7 \times 10^{-2}$ and the learning rate for all parameters decreases by the factor discussed above).

\subsubsection{Application of Dropout Methods}

The following table shows which Dropout methods were applied in each experiment, and at which
parts of the model architecture Dropout was applied. In all experiments a Dropout probability
$p = 0.2$ was used.

\begin{table}[]
\centering
    \begin{tabular}{l|c|c|c}
    \hline
    Experiment Name           & ResNet Blocks & SPP Output & Decoder Output \\
    \hline
    \texttt{none}             & None          & None       & None           \\
    \texttt{resnet-chandrop}  & Channel       & None       & None           \\
    \texttt{spp-chandrop}     & None          & Channel    & None           \\
    \texttt{decoder-chandrop} & None          & None       & Channel        \\
    \texttt{upper-chandrop}   & None          & Channel    & Channel        \\
    \texttt{all-chandrop}     & Channel       & Channel    & Channel        \\
    \texttt{resnet-uout}      & UOut Channel  & None       & None           \\
    \texttt{spp-uout}         & None          & UOut Channel & None         \\
    \texttt{decoder-uout}     & None          & None       & UOut Channel   \\
    \texttt{upper-uout}       & None          & UOut Channel & UOut Channel \\
    \texttt{all-uout}         & UOut Channel  & UOut Channel & UOut Channel \\
    \texttt{resnet-dropblock} & DropBlock         & None       & None           \\
    \texttt{spp-dropblock}    & None          & DropBlock      & None           \\
    \texttt{decoder-dropblock}& Channel       & None       & DropBlock          \\
    \texttt{upper-dropblock}  & None          & DropBlock      & DropBlock          \\
    \texttt{all-dropblock}    & DropBlock         & DropBlock      & DropBlock          \\
    \end{tabular}
    \captionof{table}{Testing Matrix for each experiment}
    \label{tab:dropout-testing-matrix}
\end{table}

\section{Experimental Results}

Experiments were performed using the aforementioned PASCAL VOC2012 dataset, by training using 10\% of the data in each class
and checking model performance on the full validation set. We report both quantitative statistics and also empirically
show the effect of the different forms of Dropout regularization on model performance on images in the validation set.

\subsection{Quantitative Results}

Empirical validation set mIoU scores and channel cross-entropy-loss scores
are reported for each experiment below:

\begin{table}[]
\centering
\begin{tabular}{l|c|c|c|c|c|c}
\hline
Experiment                & Mean mIoU  & Std.      & Worst mIoU & Med. mIoU   & Best mIoU & Loss \\
\hline
\texttt{pretrained-base}   & 0.81        & 0.09             & 0.55        & 0.82        & 0.99        & 0.02      \\
\hline
\texttt{none}              & 0.49        & 0.10             & 0.26        & 0.50        & 0.65        & 0.06      \\
\texttt{resnet-chandrop}   & \emph{0.56} & \emph{0.09}      & \emph{0.35} & 0.50        & \emph{0.77} & \emph{0.05}      \\
\texttt{spp-chandrop}      & 0.54        & 0.10             & 0.28        & 0.53        & 0.74        & 0.06      \\
\texttt{decoder-chandrop}  & 0.50        & \emph{0.09}      & 0.32        & 0.53        & 0.67        & 0.06      \\
\texttt{upper-chandrop}    & 0.49        & \emph{0.09}      & 0.28        & 0.50        & 0.68        & 0.07      \\
\texttt{all-chandrop}      & 0.55        & 0.11             & 0.27        & 0.55        & 0.76        & 0.06      \\
\texttt{resnet-dropblock}  & 0.45        & 0.10             & 0.21        & 0.45        & 0.74        & 0.08      \\
\texttt{spp-dropblock}     & 0.47        & \emph{0.09}      & 0.29        & 0.46        & 0.72        & 0.06      \\
\texttt{decoder-dropblock} & 0.51        & \emph{0.09}      & 0.34        & 0.50        & 0.68        & 0.07      \\
\texttt{upper-dropblock}   & 0.51        & 0.10             & 0.30        & 0.51        & \emph{0.77} & 0.07      \\
\texttt{all-dropblock}     & 0.48        & 0.10             & 0.28        & 0.47        & 0.69        & 0.07      \\
\texttt{resnet-uout}       & 0.50        & 0.11             & 0.31        & 0.49        & 0.74        & 0.06      \\
\texttt{spp-uout}          & 0.50        & 0.11             & 0.29        & 0.50        & 0.74        & 0.06      \\
\texttt{decoder-uout}      & 0.48        & 0.10             & 0.31        & 0.45        & 0.72        & 0.07      \\
\texttt{upper-uout}        & \emph{0.56} & \emph{0.09}      & 0.31        & \emph{0.57} & 0.72        & \emph{0.05}      \\
\texttt{all-uout}          & 0.52        & 0.10             & 0.31        & 0.51        & 0.76        & 0.06      \\
\end{tabular}
\captionof{table}{mIoU and Loss metrics for experiments. Statistics were taken from the best epoch. Best results in each category italicized}
\label{tab:metrics}
\end{table}

\begin{table}[]
\centering
\begin{tabular}{l|c|c|c|c|c|c}
\hline
Experiment                & Mean mIoU             & Std.                 & Worst mIoU          & Med. mIoU             & Best mIoU                    & Loss                 \\
\hline
\texttt{pretrained-base}   & 0.81                 & 0.09                 & 0.55                 & 0.82                 & 0.99                         & 0.02                 \\
\hline
\texttt{none}              & 0.49                 & 0.10                 & 0.26                 & 0.50                 & 0.65                         & 0.06                 \\
\texttt{resnet-chandrop}   & 0.53                 & 0.11                 & 0.30                 & 0.50                 & 0.54                         & 0.06                 \\
\texttt{spp-chandrop}      & 0.50                 & 0.10                 & 0.28                 & 0.51                 & \textbf{0.75}                & 0.06                 \\
\texttt{decoder-chandrop}  & 0.47                 & \emph{0.09}          & 0.28                 & 0.48                 & \textbf{0.68}                & 0.07                 \\
\texttt{upper-chandrop}    & \textbf{0.53}        & 0.10                 & \textbf{0.34}        & \textbf{0.53}        & \textbf{0.74}                & 0.06                 \\
\texttt{all-chandrop}      & \textbf{\emph{0.59}} & \textbf{\emph{0.09}} & \textbf{\emph{0.39}} & \textbf{\emph{0.59}} & \textbf{0.78}                & \textbf{\emph{0.05}} \\
\texttt{resnet-dropblock}  & \textbf{0.51}        & 0.10                 & \textbf{0.28}        & \textbf{0.52}        & 0.69                         & \textbf{0.07}        \\
\texttt{spp-dropblock}     & \textbf{0.52}        & 0.11                 & \textbf{0.31}        & \textbf{0.52}        & \textbf{0.75}                & 0.06                 \\
\texttt{decoder-dropblock} & \textbf{0.55}        & 0.11                 & 0.32                 & \textbf{0.56}        & \textbf{0.73}                & \textbf{0.05}        \\
\texttt{upper-dropblock}   & \textbf{0.53}        & 0.13                 & 0.29                 & \textbf{0.57}        & 0.74                         & \textbf{0.06}        \\
\texttt{all-dropblock}     & \textbf{0.49}        & 0.10                 & 0.28                 & 0.47                 & 0.69                         & 0.07                 \\
\texttt{resnet-uout}       & 0.50                 & 0.11                 & 0.31                 & 0.49                 & 0.74                         & 0.06                 \\
\texttt{spp-uout}          & 0.50                 & \textbf{0.10}        & 0.28                 & \textbf{0.51}        & \textbf{0.75}                & 0.06                 \\
\texttt{decoder-uout}      & \textbf{0.50}        & 0.10                 & \textbf{0.34}        & \textbf{0.49}        & \textbf{0.78}                & 0.07                 \\
\texttt{upper-uout}        & 0.52                 & 0.12                 & 0.23                 & 0.54                 & \textbf{\emph{0.79}}         & 0.06                 \\
\texttt{all-uout}          & 0.51                 & 0.11                 & 0.29                 & \textbf{0.52}        & 0.76                         & 0.06                 \\
\end{tabular}
\captionof{table}{mIoU and Loss metrics for experiments using \texttt{ScheduledDropPath} with \texttt{n = 30}. Statistics were taken from the best epoch. Best results in each category italicized. Results that were better than corresponding results without scheduling bolded.}
\label{tab:metrics-drop-path}
\end{table}

A few things stand out from these empirical results. First, usage of appropriate Dropout is effective
in limited data situations for improving model generalization. Second, application of dropout at the later layers only improves model generalization marginally compared to application of dropout at the earlier layers. Third, using a form of dropout that does not increase noise variance that much is better. Introducing too much variance through the use of \texttt{DropBlock} in the earlier layers causes shifts in classification as we will see later on, leading to a drop in mIoU performance. This phenomenon is better explained by Li, Xiang et al., who show that
where dropout is applied \emph{before} Batch Normalization layers, then at test time, performance will drop due to a shift in batch variance that the Batch Normalization layers were relying on. Those authors proposed \texttt{UOut} as a
mechanism for dealing with this problem, which is also trialled in these experiments.  Finally we can observe from these experimental results that when Dropout applied only in many places, the variance in mIoU scores on the validation set increases, however when Dropout is applied in one  places, total variance in scores is reduced, leading to models which may perform worse, but more consistently.

Finally, from table \ref{tab:metrics-drop-path} we observe that \texttt{ScheduledDropPath} improves mIoU performance
mainly in cases where \texttt{DropBlock} is used \cite{conf/nips/GhiasiLL18}, but it also gives a fairly large performance boost when used in
conjunction with \texttt{SpatialDropout} on both the feature detection, pooling and decoder layers of the model. Therefore, we
can say that it is beneficial to use this technique in cases where more variance is being introduced in the image signal;
it is better to allow the model to fit the signal data first, then force it to focus on other parts of the signal
in the image channels.

\newcommand{\trainingCurves}[3] {
\begin{figure}
    \centering
    \begin{subfigure}[b]{0.45\textwidth}
        \centering
        \includegraphics[width=\textwidth,height=5cm,keepaspectratio]{resources/#1/#2/train_loss.png}
        \caption[]%
        {{\small Train. Loss}}    
        \label{fig:#1-#2-train-loss}
    \end{subfigure}
    \begin{subfigure}[b]{0.45\textwidth}  
        \centering 
        \includegraphics[width=\textwidth,height=5cm,keepaspectratio]{resources/#1/#2/val_loss.png}
        \caption[]%
        {{\small Val. Loss}}    
        \label{fig:#1-#2-val-loss}
    \end{subfigure}
    \begin{subfigure}[b]{0.45\textwidth}   
        \centering 
        \includegraphics[width=\textwidth,height=5cm,keepaspectratio]{resources/#1/#2/train_mious.png}
        \caption[]%
        {{\small Train. mIoU}}    
        \label{fig:#1-#2-train-miou}
    \end{subfigure}
    \begin{subfigure}[b]{0.45\textwidth}   
        \centering 
        \includegraphics[width=\textwidth,height=5cm,keepaspectratio]{resources/#1/#2/val_mious.png}
        \caption[]%
        {{\small Val. mIoU}}    
        \label{fig:#1-#2-val-miou}
    \end{subfigure}
    \captionof{figure}{Training statistics (#3). Summaries are of best batch.} 
    \label{fig:training-stats-#1-#2}
\end{figure}
}

\newcommand{\visualizeSegmentation}[3] {
\begin{figure}
    \centering
    \begin{subfigure}[b]{0.45\textwidth}
        \centering
        \includegraphics[clip,width=\textwidth,height=15cm,keepaspectratio]{resources/#1/#2/validation_seg1_final.png}
        \caption[]%
        {{\small Segmentation for Image 1}}    
        \label{fig:#1-#2-val-seg0}
    \end{subfigure}
    \begin{subfigure}[b]{0.45\textwidth}  
        \centering 
        \includegraphics[clip,width=\textwidth,height=15cm,keepaspectratio]{resources/#1/#2/validation_seg2_final.png}
        \caption[]%
        {{\small Segmentation for Image 2}}    
        \label{fig:#1-#2-val-seg1}
    \end{subfigure}
    \captionof{figure}{Segmentation results over each epoch (#3).} 
    \label{fig:seg-results-#1-#2}
\end{figure}
}

\newcommand{\visualizeAllSegmentation}[3] {
\begin{figure}
    \centering
    \includegraphics[clip,width=\textwidth,height=15cm,keepaspectratio]{resources/#1/#2/best_images.png}
    \captionof{figure}{Best and worst segmentation results over each epoch (#3).} 
    \label{fig:seg-best-worst-#1-#2}
\end{figure}
}

\subsection{Qualitative Evaluation}
\label{sec:qualitative-evaluation}

For each experimental method, we also qualitatively evaluate the segmentations produced by
the model during the training process and once it is fitted. The same two images, one of
class \texttt{tvmonitor} and another of class \texttt{cow} were used during the qualitative
evaluation of each model.

The author notes that, intuitively, Image 1 is likely to be difficult for most model architectures to
correctly fit, due to the lack of sharp image gradients likely to be detected by convolutional
filters, especially within the inner portion of the image. While Image 2 seems easier to classify,
we observed that many architectures, even a pre-trained architecture on the complete dataset appear
to classify it incorrectly.

\subsubsection{Experiment \texttt{pretrained-base} (Figure \ref{fig:seg-results-main-pretrained})}
\label{sec:exp-pretrained-base}

\visualizeSegmentation{main}{pretrained}{\texttt{pretrained-base}}

As a baseline, we compare the segmentations produced by the reference model on PyTorch Hub, trained the subset of images in the training set for COCO2017 constituting the 20 classes in the PASCAL VOC2012 dataset. This model was trained without any regularization and for five epochs, with a batch size of 6.

We observe that with a large amount of available training data, the model
is able to segment the low-gradient areas of the image with a relatively
high degree of precision, though it does over-estimate the bounds of
the object in Image 1 to parts that are related to the object (the cabling
near the frame of the monitor). Image 2 is correctly segmented but misclassified.

\visualizeAllSegmentation{main}{pretrained}{\texttt{pretrained-base}}

By showing the best and worst segmentations produced by the pretrained model, we can
see in general that this architecture is unable to produce segmentations for objects
in unusual contexts (where the context contains a large amount of higher-frequency
components), nor is it able to perform well when finer image details. However, it is
able to generalize well where the object is clearly in the foreground and differentiated
from other low-frequency components.

\subsubsection{Experiment \texttt{none} (Figure \ref{fig:seg-results-limited-none})}
\label{sec:exp-none}

\visualizeSegmentation{limited}{none}{\texttt{none}}

Without any regularization, the model overfits to classifying most pixels as
\texttt{background} (no visual overlay). This is most apparent in
Image 1, where only the top-left corner of the image is classified as \texttt{tvmonitor}.
We also observe that while the model is able to correctly segment some features of
the cow in Image 2, many of the features are missing from the segmentation
and the correctly segmented parts are misclassified. We will see that finding the correct classification for Image
2 is a difficult problem when evaluating other models.

\visualizeAllSegmentation{limited}{none}{\texttt{none}}

We also include a visualization of the best and worst segmentations produced
by the model. The best segmentations are those where there are very sharp gradients
between the object and background and a clear difference in features and colors. On the other
hand, poor segmentations occur on images where the image makes up the background, has
similar colors and textures to the background or has macro-level features that are unlikely
to cause activations in the feature detection layers.

\subsubsection{Experiment \texttt{resnet-chandrop} (Figure \ref{fig:seg-results-limited-feature-detection-use-channel-dropout})}
\label{sec:exp-resnet-chandrop}

\visualizeSegmentation{limited}{feature-detection-use-channel-dropout}{\texttt{resnet-chandrop}}

Applying channel dropout in the feature detection layers during training changes
the segmentations produced on the validation set. In Image 1, a problem similar to the
lack of overall feature detection manifests, however the model is able to some distinguish
regions with low gradients as part of the \texttt{tvmonitor} class, as long as they are
"assisted" by the presence of sharp gradients on the bezel of the monitor. In contrast to
the \texttt{none} experiment, the model correctly is able to determine more pixels
as belonging to the \texttt{cow} class, but still misses many features.

\subsubsection{Experiment \texttt{spp-chandrop}}
\label{sec:exp-spp-chandrop}

\visualizeSegmentation{limited}{pyramid-use-channel-dropout}{\texttt{spp-chandrop} (Figure \ref{fig:seg-results-limited-pyramid-use-channel-dropout})}

On Image 1, usage of Channel Dropout on the Spatial Pyramid Pooling Layers seems very
promising; the model is able to detect more of the image segment even in regions
with low gradients. This seems to indicate that Channel Dropout at this layer
changes the optimization path such that the model better takes into account feature
maps at larger scales, where the gradients are likely to be steeper. On Image 2,
we shift into misclassification, even though we are able to detect more foreground
features. This indicates that over-reliance on features that cause classification
uncertainty and misclassification is a problem that still exists at the feature detection stage.

\subsubsection{Experiment \texttt{decoder-chandrop} (Figure \ref{fig:seg-results-limited-decoder-use-channel-dropout})}
\label{sec:exp-decoder-chandrop}

\visualizeSegmentation{limited}{decoder-use-channel-dropout}{\texttt{decoder-chandrop}}

Applying Channel Dropout at the Decoder layer assists in the detecting more of the features of the
object in the foreground of the image, though we observe further drift into misclassification, perhaps
indicating the Channel Dropout was able to improve the edge detection process but only for
certain classes such as \texttt{bus} and \texttt{motorbike} and not \texttt{cow}.

\subsubsection{Experiment \texttt{upper-chandrop} (Figure \ref{fig:seg-results-limited-everything})}
\label{sec:exp-upper-chandrop}

\visualizeSegmentation{limited}{everything}{\texttt{upper-chandrop}}

Similar to \ref{sec:exp-spp-chandrop}, we observe better marginally performance on Image 1, on sections that
contain smooth gradients.  There is still a degree of feature detection uncertainty for Image 2, as only
patches of it are detected as part of the foreground, possibly because the model
does not give a high amount of weight to features at larger image scales.

\subsubsection{Experiment \texttt{all-chandrop} (Figure \ref{fig:seg-results-limited-everything-including-feature-detection-dropout})}
\label{sec:exp-all-chandrop}

\visualizeSegmentation{limited}{everything-including-feature-detection-dropout}{\texttt{all-chandrop}}

Applying dropout on both the feature detection and upper layers of the model during
training enables the model to perform much better on Image 1, with the highest mIoU score
seen so far and a better detection of finer image components such as the corners of the
display (even though in regions where there is a low gradient in the downsampled image space between the edge of the display and the background, the model "misses" the edge
and classification trails off into the background). However, the misclassification issue
in Image 2 returns, even if finer features are able to be detected.

\subsubsection{Experiment \texttt{resnet-dropblock} (Figure \ref{fig:seg-results-limited-feature-detection-use-activation-dropout})}
\label{sec:exp-resnet-dropblock}

\visualizeSegmentation{limited}{feature-detection-use-activation-dropout}{\texttt{resnet-dropblock}}

Applying \texttt{DropBlock} on each Block of the feature detection layers appears to improve
performance in Image 1, along both the contours and partially within areas that have low gradients. However
it does so in an inconsistent manner, indicating that activations are not regularized within all
areas of the image. \texttt{DropBlock} improves feature detection in Image 2, although more complex features such
as the face of the cow are not detected.

\subsubsection{Experiment \texttt{spp-dropblock} (Figure \ref{fig:seg-results-limited-pyramid-use-activation-dropout})}
\label{sec:exp-pyramid-dropblock}

\visualizeSegmentation{limited}{pyramid-use-activation-dropout}{\texttt{spp-dropblock}}

Applying \texttt{DropBlock} on each on the pyramid pooling layers has a similar effect to \ref{fig:seg-results-limited-everything}, in that both high and low-gradient areas of Image 1 are detected, but
where contours are thin, these gradients in the image are not detected and misclassification continues into the
background. This probably indicates that finer edge detection was not properly trained for features of
this scale and orientation. \texttt{DropBlock} improves feature detection in Image 2, although more
complex features such as the face of the cow are not detected.

\subsubsection{Experiment \texttt{decoder-dropblock} (Figure \ref{fig:seg-results-limited-decoder-use-activation-dropout})}
\label{sec:exp-decoder-dropblock}

\visualizeSegmentation{limited}{decoder-use-activation-dropout}{\texttt{decoder-dropblock}}

Applying \texttt{DropBlock} on the decoder layers causes large amounts of misclassification on certain features on Image 1, leaving only a few patches where
features are distinctive changes in image gradients (such as hard corners between
screen and the bezels) are correctly detected and classified. The rest is misclassified as \texttt{train}. On Image 2, misclassification problems also occurr,
through the model is able to better identify complex features such as the face
of the cow as being part of the foreground.

\subsubsection{Experiment \texttt{all-dropblock} (Figure \ref{fig:seg-results-limited-everything-including-feature-detection-activation-dropout})}
\label{sec:exp-all-dropblock}

\visualizeSegmentation{limited}{everything-including-feature-detection-activation-dropout}{\texttt{all-dropblock}}

Applying \texttt{DropBlock} over all layers does not have desirable effects, both images have the worst
mIoU score by far. Especially for Image 1, \texttt{DropBlock} regularization on all layers including the feature
detection layers, breaks almost all feature detection, even where gradients are sharp. On Image 2, feature detection
is retained, though the dominant class shifts and the foreground is misclassified.

\subsubsection{Experiment \texttt{resnet-uout} (Figure \ref{fig:seg-results-limited-feature-detection-use-channel-uout})}
\label{sec:exp-resnet-uout}

\visualizeSegmentation{limited}{feature-detection-use-channel-uout}{\texttt{resnet-uout}}

Applying \texttt{UOut} just to the feature detection layers also decreases performance, causing very low precision on the image 1 and misclassification
on Image 2.

\subsubsection{Experiment \texttt{spp-uout} (Figure \ref{fig:seg-results-limited-pyramid-use-channel-uout})}
\label{sec:exp-pyramid-uout}

\visualizeSegmentation{limited}{pyramid-use-channel-uout}{\texttt{spp-uout}}

Applying \texttt{UOut} on each of the pyramid pooling layers improves the detection of high-gradient regions, but does not improve the detection of the
matte low-gradient textures on Image 1. On Image 2, \texttt{UOut} improves feature detection such that the face of the cow is now detected and it
does not suffer from the misclassification issues seen in \texttt{DropBlock} and \texttt{SpatialDropout}.

\subsubsection{Experiment \texttt{decoder-uout} (Figure \ref{fig:seg-results-limited-decoder-use-channel-uout})}
\label{sec:exp-decoder-uout}

\visualizeSegmentation{limited}{decoder-use-channel-uout}{\texttt{decoder-uout}}

Applying \texttt{UOut} on the decoder layers does not appear to help all that much, in the same way that applying \texttt{DropBlock} causes poor foreground detection performance. On Image 1, only parts of the small parts of the foreground are detected and on Image 2, most of the object is misclassified, though the face of the cow is correctly classified.

\subsubsection{Experiment \texttt{upper-uout} (Figure \ref{fig:seg-results-limited-everything-uout})}
\label{sec:exp-upper-uout}

\visualizeSegmentation{limited}{everything-uout}{\texttt{upper-uout}}

Applying \texttt{UOut} over the upper layers enables detection of the entire foregrund in Image 1, though a model trained in this way has low precision and classifies past the edges of foreground object. This is perhaps because \texttt{UOut} breaks detection of edge detectors for this object type at the given orientations. We also see decent forground detection performance on Image 2, though the foreground object is again misclassified.

\subsubsection{Experiment \texttt{all-uout} (Figure \ref{fig:seg-results-limited-everything-including-feature-detection-uout})}
\label{sec:exp-all-uout}

\visualizeSegmentation{limited}{everything-including-feature-detection-uout}{\texttt{all-uout}}

Applying \texttt{UOut} over all layers exhibits similar problems as seen in \ref{sec:exp-upper-uout} and \ref{sec:exp-upper-uout}. Low precision is observed
on the left hand side of Image 1 and misclassification is observed on Image 2.

\section{Conclusions}

This work contributes thorough experimentation showing that correctly motivated Dropout algorithms
applied to image Segmentation models in low-data scenarios can be an effective mechanism for improving
segmentation performance. We show that the most important dimension upon which to apply Dropout is along
the channels - dropping patches of the image using the \texttt{DropBlock} algorithm yield useful results,
but is not sufficient to prevent overfitting to features activated by noisy image gradients. The effectiveness
of Channel Dropout also suggests that different image features are indeed detected in different channels.

We also show through exploration of segmentation results over training that Channel Dropout when applied
to different layers of \texttt{DeepLabV3+} at training time. For instance, we show that the channels
at the feature detection layers (the ResNet backbone) appear to be most influential in determining
the most likely class of the image, from which the class of the segmentation is determined. We also
show that the Spatial Pooling Pyramid layers can benefit from channel regularization, in that they are
better able to produce activations taking into account all scales of the image, including the smaller
scales; as opposed to just the larger scales which encode more fine-grained noisy image gradients.

Finally, we show that applying a linear schedule to dropout probability can be useful in cases where information loss is applied to the within the signal itself, as opposed to the signal channels. We find that a non-negligible improvement in validation mIoU over a non-scheduling baseline is achieved when this technique is used in conjunction with \texttt{DropBlock}.

\bibliographystyle{unsrt}

\begin{thebibliography}{10}

\bibitem{journals/corr/HintonVD15}
Geoffrey~E. Hinton, Oriol Vinyals, and Jeffrey Dean.
\newblock Distilling the knowledge in a neural network.
\newblock {\em CoRR}, abs/1503.02531, 2015.

\bibitem{conf/nips/GhiasiLL18}
Golnaz Ghiasi, Tsung-Yi Lin, and Quoc~V. Le.
\newblock Dropblock: A regularization method for convolutional networks.
\newblock In Samy Bengio, Hanna~M. Wallach, Hugo Larochelle, Kristen Grauman,
  Nicolò Cesa-Bianchi, and Roman Garnett, editors, {\em NeurIPS}, pages
  10750--10760, 2018.

\bibitem{conf/cvpr/TompsonGJLB15}
Jonathan Tompson, Ross Goroshin, Arjun Jain, Yann LeCun, and Christoph Bregler.
\newblock Efficient object localization using convolutional networks.
\newblock In {\em CVPR}, pages 648--656. IEEE Computer Society, 2015.

\bibitem{conf/icml/IoffeS15}
Sergey Ioffe and Christian Szegedy.
\newblock Batch normalization: Accelerating deep network training by reducing
  internal covariate shift.
\newblock In Francis~R. Bach and David~M. Blei, editors, {\em ICML}, volume~37
  of {\em JMLR Workshop and Conference Proceedings}, pages 448--456. JMLR.org,
  2015.

\bibitem{journals/corr/abs-1801-05134}
Xiang Li, Shuo Chen, Xiaolin Hu, and Jian Yang.
\newblock Understanding the disharmony between dropout and batch normalization
  by variance shift.
\newblock {\em CoRR}, abs/1801.05134, 2018.

\bibitem{journals/corr/abs-1802-02611}
Liang-Chieh Chen, Yukun Zhu, George Papandreou, Florian Schroff, and Hartwig
  Adam.
\newblock Encoder-decoder with atrous separable convolution for semantic image
  segmentation.
\newblock {\em CoRR}, abs/1802.02611, 2018.

\bibitem{journals/corr/HeZRS15}
Kaiming He, Xiangyu Zhang, Shaoqing Ren, and Jian Sun.
\newblock Deep residual learning for image recognition.
\newblock {\em CoRR}, abs/1512.03385, 2015.

\bibitem{Chol17Xception}
Fran{\c{c}}ois Chollet.
\newblock Xception: Deep learning with depthwise separable convolutions.
\newblock In {\em Proceedings of the IEEE conference on computer vision and
  pattern recognition}, pages 1251--1258, 2017.

\bibitem{journals/corr/HeZR014}
Kaiming He, Xiangyu Zhang, Shaoqing Ren, and Jian Sun.
\newblock Spatial pyramid pooling in deep convolutional networks for visual
  recognition.
\newblock {\em CoRR}, abs/1406.4729, 2014.

\bibitem{srivastava2014dropout}
Nitish Srivastava, Geoffrey~E Hinton, Alex Krizhevsky, Ilya Sutskever, and
  Ruslan Salakhutdinov.
\newblock Dropout: a simple way to prevent neural networks from overfitting.
\newblock {\em Journal of Machine Learning Research}, 15(1):1929--1958, 2014.

\bibitem{zhang2019shiftinvar}
Richard Zhang.
\newblock Making convolutional networks shift-invariant again.
\newblock In {\em ICML}, 2019.

\bibitem{conf/cvpr/ZophVSL18}
Barret Zoph, Vijay Vasudevan, Jonathon Shlens, and Quoc~V. Le.
\newblock Learning transferable architectures for scalable image recognition.
\newblock In {\em CVPR}, pages 8697--8710. IEEE Computer Society, 2018.

\bibitem{pascal-voc-2012}
M.~Everingham, L.~Van~Gool, C.~K.~I. Williams, J.~Winn, and A.~Zisserman.
\newblock The {PASCAL} {V}isual {O}bject {C}lasses {C}hallenge 2012 {(VOC2012)}
  {R}esults.
\newblock
  http://www.pascal-network.org/challenges/VOC/voc2012/workshop/index.html,
  2012.

\bibitem{jfzhang-repo}
Jianfeng Zhang.
\newblock Deeplab v3+ in pytorch, 2018.

\bibitem{ChePap17}
Liang-Chieh Chen, George Papandreou, Iasonas Kokkinos, Kevin Murphy, and
  Alan~L. Yuille.
\newblock Deeplab: Semantic image segmentation with deep convolutional nets,
  atrous convolution, and fully connected crfs.
\newblock {\em CoRR}, abs/1606.00915, 2016.

\end{thebibliography}

\end{document}